\documentclass{article}

\usepackage{arxiv}

\usepackage{url}
\usepackage{listings}
\usepackage{cite}
\usepackage{amsmath,amssymb,amsfonts}
\usepackage{algorithmicx}
\usepackage{algorithm}
\usepackage{algpseudocode}
\usepackage{graphicx}
\usepackage{booktabs}
\usepackage{float}
\usepackage{hyperref}
\usepackage{multicol}
\usepackage{enumitem}

\usepackage{textcomp}
\usepackage{multirow}
\usepackage{xcolor}
\usepackage{adjustbox}
\usepackage[english]{babel}

\usepackage[utf8]{inputenc}
\usepackage[T1]{fontenc}
\usepackage{tabularx}

\newcommand{\comment}[1]{}
\def\BibTeX{{\rm B\kern-.05em{\sc i\kern-.025em b}\kern-.08em
    T\kern-.1667em\lower.7ex\hbox{E}\kern-.125emX}}

\usepackage{color}

\newcommand{\entra}[2][]{%
   {\color{blue}#2}%
   \ifx&#1&%
      {}
    \else%
      \footnote{{\color{blue}(entra):} #1}
    \fi%
}%

\newcommand{\sai}[2][]{%
   {\color{black}\sout{#2}}%
   \ifx&#1&%
      {}
    \else%
      \footnote{{\color{black}(sai):} #1}
    \fi%
}%

\newcommand{\rever}[2][]{%
   {\color{magenta}#2}%
   \ifx&#1&%
      {}
    \else%
      \footnote{{\color{magenta}(rever):} #1}
    \fi%
}%

\newcommand{\todo}[2][]{%
   {\color{red}[\textbf{TODO}: \textit{#2}]}%
   \ifx&#1&%
      {}
    \else%
      \footnote{{\color{red}[(todo):} #1]}
    \fi%
}%

\begin{document}

\title{\textbf{Cabrita: closing the gap for foreign languages}}

\author{Celio Larcher \and Marcos Piau \and Paulo Finardi \and Pedro Gengo \and Piero Esposito \and Vinicius Caridá}

\date{\texttt{email: \{celiolarcher, marcos.piau.vieira, pfinardi, pedro.gengo.lourenco, piero.skywalker, vfcarida\}@gmail.com} \vspace{0.4cm} }

\maketitle
\begin{abstract}
The strategy of training the model from scratch in a specific language or domain serves two essential purposes: i) enhancing performance in the particular linguistic or domain context, and ii) ensuring effective tokenization. The main limitation inherent to this approach lies in the associated cost, which can reach six to seven-digit dollar values, depending on the model size and the number of parameters involved.

The main solution to overcome the cost challenge is to rely on available pre-trained models, which, despite recent advancements such as the LLaMA and LLaMA-2 models, still demonstrate inefficiency for certain specific domain problems or prove ineffective in scenarios involving conversational memory resources, given the large number of tokens required to represent text.

To overcome this issue, we present a methodology named Cabrita, which, as our research demonstrates, successfully addresses the performance and efficient tokenization problem, all at an affordable cost. We believe that this methodology can be applied to any transformer-like architecture model. To validate the study, we conducted continuous pre-training exclusively using Portuguese text on a 3-billion-parameter model known as OpenLLaMA, resulting in a model named openCabrita 3B. The openCabrita 3B also features a new tokenizer that results in a significant reduction in the number of tokens required to represent the text. In our assessment, for few-shot learning tasks, we achieved similar results with this 3B model compared to a traditional continuous pre-training approach as well as to 7B models English pre-trained models.
\end{abstract}

\section{Introduction}
While the Portuguese language does not suffer from a lack of data for training transformer-based language models, a native speaker can readily perceive limitations in text generation and performance of pre-trained models predominantly based on English language data. Following the research conducted in Sabiá \cite{sabia}, we share the perspective that the conventional practice of simultaneously pre-training models in multiple languages fails to capture the cultural richness and intrinsic knowledge of each language.

As demonstrated in the study \cite{HowGoodisYourTokenizer}, language-specific adapted tokenizers have the potential to significantly enhance the monolingual performance of multilingual models.

Considering that pre-training a model in a single language with a specific tokenizer leads to notably superior performance, it's natural to question why there aren't more language-specific models. However, the answer lies in a crucial factor: cost. An example of this is the cost of the LLaMa 2 7B model, which required a total of 184,320 hours of processing on GPU A-100 80-GB. Assuming a conservative estimate that 1 hour of this GPU costs \$1, the resulting cost approaches \$200,000 significantly. This underscores how the substantial investments required to create and maintain language-specific language models can pose a significant financial challenge.

This article investigates an alternative approach to zero-shot pre-training based on three fundamental principles: low cost, performance optimization, and efficient tokenization. This approach aims to enable the model to use fewer tokens and minimize inference time compared to existing multilingual models.


\section{Methodology}

\subsection{LLaMA Models}

The LLaMA models are Large Language Models developed by the Meta AI's team \cite{touvron2023llama}. Based on the well-established transformers architecture it incorporates several novel training and inference mechanisms, notably the use of Pre-normalization \cite{zhang2019root}, the SwiGlu activation function \cite{shazeer2020glu} and the inclusion of Rotary Embeddings \cite{su2022roformer}.

The primary distinction of LLaMA models lies in the magnitude of pre-training data utilized during their development, scaling from the standard 300 billion tokens in the optimal Chinchila's law pattern \cite{hoffmann2022training} to an extensive 1 trillion tokens. Although this setup incurs increased resource costs during training, it significantly improves the state-of-the-art (SOTA) performance for all model sizes at the time of release. As a result, inference costs are reduced, as smaller models can achieve comparable levels of performance, thereby justifying the theoretical non-optimal behavior during pre-training.

After its introduction, several other Large Language Models have been trained following a similar approach, facilitating further advancements in SOTA standards, particularly for smaller models. For this study, we adopt the OpenLLaMA implementation\cite{openlm2023openllama}, an open-source Apache 2.0 alternative with the same recipe as the open-science-only LLaMA.

\subsection{openCabrita Models}

Despite the proliferation of numerous open source Large Language Models in recent years, the availability of generative models for non-English languages remains a persistent problem.

While employing English models for other languages might seem feasible, their performance typically falls short of optimal, even when utilizing multilingual options or implementing Instruction Fine Tuning \cite{longpre2023flan} in the target language \cite{bloomz, llama_chines, sabia}.

Taking this aspect into consideration, this work introduces the openCabrita models, a collection of checkpoints derived from the OpenLLama models. These models undergo additional pre-training using a Portuguese corpus, inspired by the methodology proposed in \cite{sabia}, while also incorporating additional tokenizer adaption. 
The training procedure will be elaborated on in the next subsections.

\subsubsection{Tokenizer adaptation}

One of the main challenges of adapting a Large Language Model to a new language that is underrepresented in its pre-training is the tokenizer behavior. 
This is the case for the Portuguese language in the OpenLLaMA model, where despite the 1 trillion tokens used, just a small part were non-English.  
The default tokenizer, in such cases, tends to be overly verbose for non-English examples, necessitating the division of the text into too small parts.

This behavior is undesirable for two primary reasons: i) it shatters the amount of information contained within each sub-unit processed by the model, potentially making it more challenging to understand long-term relations within the context; and ii) it escalates the computational cost of both training and inference, as it increases the number of tokens that need to be processed for the same text sample.

However, the inclusion of a new tokenizer in a pre-trained model poses a challenge, as there exists a direct mapping between the tokens used in pre-training and the model's embedding representation. Merely replacing the tokenizer would break this mapping, likely resulting in odd model behavior and potentially requiring a considerable amount of additional pre-training to learn the new association pattern.

To address this issue, the approach taken in this work involves adapting the tokenizer to the new language while preserving the knowledge of the original language already embedded in the model. Inspired by the work of \cite{llama_chines}, this process involved three steps:

\begin{itemize}
    \item Training a new tokenizer exclusively on Portuguese data sourced from Wikipedia. SentencePiece \cite{kudo-richardson-2018-sentencepiece} implementation was used for this purpose, resulting in a tokenizer with 40,000 tokens.
    \item Merging the original tokenizer with the newly created Portuguese tokenizer. During this merging process, all tokens present in the original tokenizer were retained, and the new Portuguese tokens that were not yet represented in the vocabulary were appended until a total of 52,000 tokens were reached. The precedence level of each merge -- BPE Score in the SentencePiece implementation -- was maintained from both tokenizers, given that this led to a better token compression level.
    \item Resizing the model's embedding and head matrix to accommodate the newly added tokens. The new rows corresponding to the Portuguese tokens were appended to the end of the original matrix to ensure that the existing token mapping remained unaffected.
\end{itemize}

The result was a bilingual tokenizer, able to handle both English and Portuguese content efficiently.

\subsubsection{Continuous pre-training}

We use the Portuguese subset of the mC4 dataset \cite{mc4} (hereafter called \texttt{mC4-pt}) as our pretraining corpus. Following the recipe in \cite{sabia}, we also apply quality filters based on MassiveText \cite{massive-text} to ensure our model is trained using high-quality documents.


First, we normalize each document independently using \texttt{ftfy} \cite{ftfy}. Then, we apply the quality filters. Unlike MassiveText, which only considers the stop words \{\textit{the, be, to, of, and, that, have, with}\}, we use the Portuguese stop words from \texttt{nltk} \cite{ntlk}; we also keep only documents with at least 200 unique tokens, following the the procedure outlined in \cite{sabia}. A document is kept in the dataset if it does not fall into any of the quality filters. Table \ref{tab:mc4_filters} shows the contribution of each filter to the final filtered dataset.


To estimate the total number of training tokens without processing the entire dataset, we assume that the remaining shards have similar sizes and behavior with respect to the quality filters. Based on this estimation, the unfiltered \texttt{mC4-pt} training dataset consists of approximately 169 million documents and 188 billion tokens, which reduces to 133 million documents and 170 billion tokens after applying the low-quality filters. To convert these documents into training examples, we add \texttt{<|bos|>} and \texttt{<|eos|>} tokens to each document, and then pack these documents into sequences of 2048 tokens.

We perform the continuous pre-training step on TPUs using the EasyLM, which is a framework built in JAX/Flax and is a one-stop solution for training LLMs on TPU and GPU devices.


To perform continued pre-training in the Portuguese language, the original script and hyper-parameters of OpenLLaMA training were employed. The model was initialized with the OpenLLaMA weights and then trained on an additional 7 billion tokens extracted from the Portuguese dataset.

The learning objective was the standard causal language modeling loss with the learning rate set to increase linearly from 0 to 3e-4 over 2,000 steps, followed by a cosine decrease to 3e-5 for the remaining 248,000 steps. The optimization algorithm used was AdamW, with a weight decay of 0.1, $\beta_1$ set to 0.9, $\beta_2$ set to 0.99, and a gradient clip value of 1.0.

For training, a TPU v3-8 was utilized, with batches of 16, each containing a sequence of 2048 tokens. To achieve the target of 2048 samples in a batch, 128 accumulation steps were performed. This configuration resulted in a throughput of 7,900  tokens/second for the 3B model.

\section{Evaluation}

\subsection{Tokenizer efficiency}

\begin{figure}[htb!]
\centering
\includegraphics[width=0.9\linewidth]{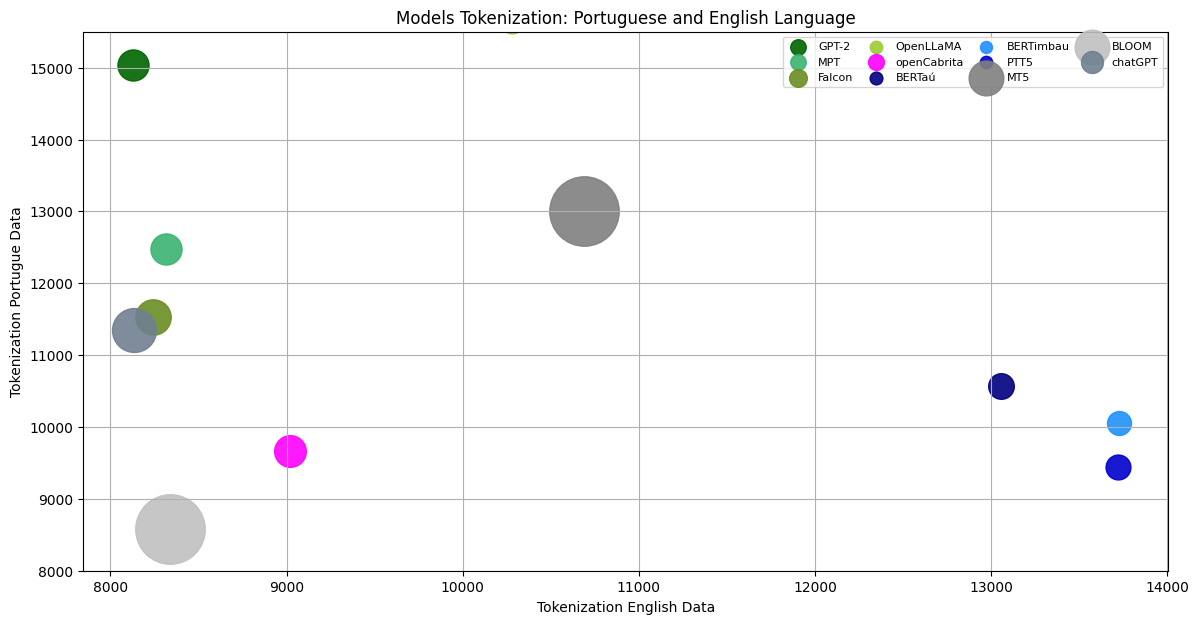} 
\caption{Tokenizer efficiency: the X axis show the number of tokens required to represent 7400 words of the Constitutional law of the USA in English, while the Y axis shows the same, but using a Portuguese translated version. The size of each sphere represents the vocabulary size of each tokenizer.}
\label{fig:tok_effic}
\end{figure}

For evaluating how alterations to the tokenizer affect text representation, a set of experiments were performed. These experiments involved the comparison of various tokenizer alternatives coming from different models and trained on distinct corpora. Figure \ref{fig:tok_effic} presents a comparison of the token count required to represent 7400 words of the Constitutional law of the USA in both English and Portuguese (translated) for each of these tokenizer variations.


It can be noted that the standard OpenLLaMA implementation demonstrates highly inefficient behavior when applied to Portuguese data. It necessitates over 50\% more tokens for representing the same text, in contrast to the multilingual Bloom and Portuguese pre-trained models  (e.g. BERTimbau \cite{bertimbau}, PTT5 \cite{ptt5}, Bertaú \cite{bertau}). While certain other LLaMA-based implementations exhibit improved token counts, such as MPT \cite{mpt} and Falcon \cite{refinedwebfalcon}, some inefficiency remains. This observation holds true even in comparison with chatGPT \cite{instructgpt}, which possesses a larger vocabulary than these LLaMA implementations.

In contrast to these choices, the Cabrita tokenizer demonstrated the ability to enhance the OpenLLaMA token requirements. It achieved a reduction of over 35\% in the token requirements, bringing it in line with the token counts of the Bloom and native Portuguese implementations.

Furthermore, upon examining the token count in English, the Cabrita tokenizer even exhibited a slight improvement in the token count when compared with the original OpenLLaMA tokenizer, with no indication of degradation in this regard.

Ultimately, the Cabrita approach presents an excellent trade-off between Portuguese and English considerations, especially when seeking models with comparable vocabulary sizes (52.000 tokens for Cabrita versus 250.680 tokens for Bloom).

\subsection{Portuguese Benchmark results}

To assess the efficacy of this proposal, 8 Portuguese evaluation datasets were employed across a variety of tasks: ASSIN 2 RTE and STS \cite{assin2}, FaQuAD \cite{faquad}, TweetSentBr \cite{tweetsentbr} are inherently in Portuguese, while for AG News \cite{agnews}, IMDB \cite{imdb}, SST2 \cite{sst2} and BoolQ \cite{boolq} translated versions provided by the authors of the Sabiá series paper were utilized.
Poeta's benchmark \cite{sabia} was not executed in our study due to the unavailability of its implementation at the time of our publication.

Table \ref{tab:dataset_info} provides an overview of the datasets along with the specific few-shot configuration employed in each experiment.

\newcolumntype{L}{>{\centering\arraybackslash}m{0.15\textwidth}}
\newcolumntype{C}{>{\centering\arraybackslash}m{0.1\textwidth}}
\newcolumntype{T}{>{\centering\arraybackslash}m{0.13\textwidth}}

\begin{table}[htbp]
\caption{Number of few-show samples and main metric for each task}
\centering\centering\resizebox{0.7\textwidth}{!}{
\begin{tabular}{lccc}
\hline
\textbf{Task} & \textbf{Type} & \textbf{Metric} & \multicolumn{1}{C}{\textbf{Few-shot Samples}} \\ \hline
FaQuAD & Extractive QA & F1 & 4 \\
ASSIN 2 RTE & Binary classification & F1-macro & 20 \\
ASSIN 2 STS & Regression & Pearson & 15 \\
TweetSentBr & Multiclass classification & F1-macro & 30 \\
SST2 & Binary classification & Accuracy & 34 \\
IMDB & Binary classification & Accuracy & 2 \\
AGNews & Multiclass classification & Accuracy & 12 \\
BoolQ & Binary classification & Accuracy & 4 \\
\hline
\end{tabular}
}
\label{tab:dataset_info}
\end{table}

A diverse array of tasks was carefully selected, spanning from binary and multi-class classification to Extractive Question Answering (QA). However, the absence of a comprehensive Portuguese benchmark prevented the utilization of a more extensive task set.


The Table \ref{tab:openllama_results} exhibits the performance comparison of openCabrita3B with the base openLLaMA3B and a subsequent pre-training version that employs the openLLaMA tokenizer (referred to as openCabrita3BPTOnly).

\begin{table}[htbp]
\caption{This Table shows the results obtained for each task and each model. The configurations are the ones presented in the Table \ref{tab:dataset_info}.}
\centering\centering\resizebox{0.8\textwidth}{!}{
\begin{tabular}{lccc}
\hline
\textbf{Task} & \textbf{open LLaMA3b} & \textbf{openCabrita3B PTOnly} & \textbf{open Cabrita3B} \\ \hline
FaQuAD & 58.71 & \textbf{66.72} & 62.16 \\
ASSIN 2 RTE & 43.38 & 38.83 & \textbf{67.36} \\
ASSIN 2 STS & \textbf{18.22} & 12.38 & 12.24 \\
TweetSentBr & 44.48 & 47.60 &\textbf{ 54.03} \\
SST2 & 86.69 & \textbf{90.25} & 89.00 \\
IMDB & 80.60 & 78.28 & \textbf{86.02} \\
AGNews & 60.39 & \textbf{67.76} & 64.98 \\
BoolQ & 61.92 & \textbf{64.09} & 63.11 \\
\hline  
\end{tabular}
}
\label{tab:openllama_results}
\end{table}

The results are mixed but somewhat favoring openCabrita3B. When openCabrita3B performs better, it is notably superior, as in ASSIN 2 RTE,  TweetSentBr, and IMDB tasks. When openCabrita3BPTOnly outperforms, openCabrita3B is usually close, just a few scores behind. While ASSIN 2 STS is the only task where the original openLLaMA3b shows a better performance, which is unexpected but consistent with what was observed in other experiments \cite{sabia}.

Nevertheless, it is evident that the Cabrita approach, which involves adapting the tokenizer, can offer at least a comparable performance level in contrast to the conventional continued pre-training. This equivalent performance improvement also comes with the added benefit of enhanced inference efficiency.

Table \ref{tab:benckmark_sabia} shows the same set of tasks, with a focus on comparing the performance of openCabrita3B to the outcomes presented in the Sabiá paper. It's important to highlight that while the datasets remain the same, possible variations in the execution scripts require a cautious approach when drawing comparisons.

\begin{table}[htbp]
\caption{This Table shows the results obtained for each task and each model for Portuguese datasets. The configurations are presented in Table \ref{tab:dataset_info}. The * symbol represents results that come from a different running script than ours, so the results need to be compared carefully.}
\centering\centering\resizebox{0.8\textwidth}{!}{
\begin{tabular}{lccccc}
\hline
\textbf{Model} & \textbf{open Cabrita3B} & \textbf{GPT-J*} & \textbf{Sabiá-J*} & \textbf{LLaMa-7B*} & \textbf{Sabiá-7B*} \\ \hline
FaQuAD & 62.16 & 59.52 & 69.28 & 77.38 & 77.43 \\
ASSIN 2 RTE & 67.36 & 54.88 & 35.49 & 56.82 & 64.87 \\
ASSIN 2 STS & 12.24 & 17.86 & 22.97 & 7.39 & 13.63 \\
TweetSentBr & 54.03 & 20.98 & 64.16 & 44.19 & 67.17 \\
SST2 & 89.00 & 83.94 & 87.16 & 88.76 & 90.69 \\
IMDB & 86.02 & 72.68 & 90.86 & 86.92 & 92.7 \\
AGNews & 64.98 & 64.15 & 84.3 & 76.94 & 83.28 \\
BoolQ & 63.11 & 48.75 & 51.53 & 57.37 & 64.07 \\
\hline
\end{tabular}
}
\label{tab:benckmark_sabia}
\end{table}

The performance of openCabrita3B seems very satisfactory for a model of its size. It consistently outperforms GPT-J, a model with 6B parameters, and demonstrates competitiveness with LLaMA-7B. In the case of continued pre-training versions of these models, openCabrita3B closely approaches the performance of Sabiá-J, while slightly lagging behind Sabiá-7B.

These results do not cast any negative light on openCabrita3B, as it performs admirably despite having only half the parameter size compared to the other four options.

\subsection{English Benchmark results}

To assess the efficacy of this approach and demonstrate its bilingual capabilities, we conducted a series of evaluations using diverse English datasets, ensuring a comprehensive understanding of its performance. Table \ref{tab:benckmark_english} presents the performance of openCabrita in comparison with models trained mainly in English and without any fine-tuning for a specific language.

Despite a slight decrease in performance for English, the model's capabilities remained robust and competitive across languages. Furthermore, we investigated the impact of different tokenizers—Open Llama and Cabrita—on the model's output. Encouragingly, both tokenizers yielded highly consistent results, indicating that the introduction of new tokens had minimal impact on the model's overall performance. This observation highlights the model's resilience and suggests its versatility for diverse applications without significant trade-offs in performance.

\begin{table}[htbp]
\caption{This Table shows the results obtained for each task and each model for English datasets. The * symbol represents results that come from a different running script than ours, so the results need to be compared carefully.}
\centering\centering\resizebox{1.\textwidth}{!}{
\begin{tabular}{lccccc}
\hline
\textbf{Task (metric)} & \textbf{open Cabrita3B} & \textbf{openCabrita3B PTOnly} & \textbf{openLLama 3B*} & \textbf{GPT-J*} & \textbf{LLaMa-7B*} \\ \hline
anli\_r1 (acc)             & 35.0 & 33.2 & 33.0 & 32.0 & 35.0 \\
anli\_r2 (acc)             & 34.0 & 33.5 & 32.0 & 34.0 & 34.0 \\
anli\_r3 (acc)             & 36.0 & 33.4 & 35.0 & 35.0 & 37.0 \\
arc\_challenge (acc)       & 28.0 & 28.1 & 34.0 & 34.0 & 39.0 \\
arc\_challenge (acc\_norm) & 32.0 & 31.9 & 37.0 & 37.0 & 41.0 \\
arc\_easy (acc)            & 59.0 & 58.0 & 69.0 & 67.0 & 62.0 \\
arc\_easy (acc\_norm)      & 54.0 & 53.6 & 65.0 & 62.0 & 52.0 \\
boolq (acc)                & 63.0 & 62.9 & 68.0 & 66.0 & 75.0 \\
hellaswag (acc)            & 41.0 & 41.1 & 49.0 & 50.0 & 56.0 \\
hellaswag (acc\_norm)      & 53.0 & 54.0 & 67.0 & 66.0 & 73.0 \\
openbookqa (acc)           & 21.0 & 20.6 & 27.0 & 29.0 & 29.0 \\
openbookqa (acc\_norm)     & 31.0 & 32.0 & 40.0 & 38.0 & 41.0 \\
piqa (acc)                 & 69.0 & 69.4 & 75.0 & 75.0 & 78.0 \\
piqa (acc\_norm)           & 69.0 & 69.5 & 76.0 & 76.0 & 78.0 \\
record (em)                & 83.0 & 83.0 & 88.0 & 88.0 & 91.0 \\
record (f1)                & 82.0 & 82.2 & 89.0 & 89.0 & 91.0 \\
rte (acc)                  & 55.0 & 52.7 & 58.0 & 54.0 & 56.0 \\
thuthfulqa\_mc (mc1)       & 25.0 & 23.1 & 22.0 & 20.0 & 21.0 \\
thuthfulqa\_mc (mc2)       & 38.0 & 37.1 & 35.0 & 36.0 & 34.0 \\
wic (acc)                  & 50.0 & 50.0 & 48.0 & 50.0 & 50.0 \\
winogrande (acc)           & 56.0 & 58.7 & 62.0 & 64.0 & 68.0 \\
\hline
\end{tabular}
}
\label{tab:benckmark_english}
\end{table}


\section{Conclusion and next steps}

This work is currently in progress and requires further development in terms of a more comprehensive set of models and experiments before arriving at any definitive conclusions. However, the outcomes obtained from modifying the tokenizer prior to performing continuous pre-training seem very promising.

Even in the absence of state-of-the-art performance enhancements, the openCabrita3b reaches a comparable metrics level when compared to the standard continuous pre-training approach while also improving inference time — an imperative factor in the context of deploying models for such applications that is commonly overlooked.

Looking ahead, the authors have outlined the following as their forthcoming initiatives:

\begin{itemize}
    \item Test the same strategy in a more diverse set of base models
    \item Establish some type of benchmark and reperform  runs for other models in order to have a comparison with the same running methodology
\end{itemize}

\section{Limitations}

\textbf{Scaling beyond 3B and for other languages} Despite our best efforts, certain ranges of model sizes and experiments remain beyond our reach due to budget constraints. Consequently, our experimentation was limited to the Portuguese language and 3B size models.

But, since the good results obtained, we are firmly convinced that employing larger-scale models could yield results at least as promising.  This line of thinking could similarly be applied to other foreign languages, particularly considering the successful experiment with Chinese presented in \cite{llama_chines}.

\textbf{Comparison to other base models and adapting strategies} Due to the absence of a structured benchmark like Harness \cite{eval-harness} for the Portuguese language, our capacity to compare with other models and strategies for the same tasks is confined to the experiments we are able to perform. Consequently, the scope of comparisons we can undertake in this paper is restricted, both in terms of quantity and depth.

\section{Acknowledgments}
We thank Google Cloud for the TPU grant through the TRC program.

\bibliographystyle{IEEEtran}
\bibliography{cabrita}

\newpage

\appendix

\section{Tokenizer Comparison}
\begin{table}[htbp]
\caption{Comparison between models tokenizers. The data is the first 7400 words of the Constitutional law of the United States which was translated for the Portuguese language. } 
\centering\centering\resizebox{0.8\textwidth}{!} {
\begin{tabular}{lrrrrr}
\hline
\textbf{Model} & \textbf{Vocab Size} & \textbf{Tokens (Portuguese data)} &  \textbf{Tokens (English data)} \\ \hline
GPT-2           & 50257 & 15036 & 8126 \\
MPT             & 50254 & 12470 & 8313 & \\
Falcon          & 65024 & 11531 & 8240 & \\
OpenLLaMA       & 32000 & 15648 & 10280 & \\
openCabrita 3B & 52000 & 9666  & 9017 & \\
BERTaú          & 34100 & 10573 & 13054 & \\
BERTimbaú       & 29794 & 10057 & 13725 & \\
PTT5            & 32100 & 9447 & 13720 & \\
MT5             & 250100 & 13010 & 10688 & \\
BLOOM           & 250680 & 8582 & 8337 & \\
chatGPT         & 100277 & 11351 & 8136 & \\
\hline
\end{tabular}
}
\label{tab:tokenizers_comparison1}
\end{table}

\begin{table}[htbp]
\caption{Comparison between tokenization strategy. The data is the first 7400 words of the Constitutional law of the United States which was translated for the Portuguese language. } 
\centering\centering\resizebox{0.8\textwidth}{!} {
\begin{tabular}{lrrrrr}
\hline
\textbf{Model} & \textbf{Vocab Size} & \textbf{Tokens (Portuguese data)} &  \textbf{Tokens (English data)} \\ \hline
openCabrita 3B & 52000 & 9666  & 9017 & \\
Cabrita 0x      & 52000 & 10169  & 9753 & \\
Cabrita 0.5x    & 52000 & 9694  & 9021 & \\
Cabrita 2x      & 52000 & 9632  & 9010 & \\
\hline
\end{tabular}
}
\label{tab:tokenizers_comparison2}
\end{table}

\section{\texttt{mC4-pt} cleaning filters}

\begin{table}[htbp]
\caption{Impact of each low-quality filter on the \texttt{mC4-pt} final clean training dataset.}
\centering\centering\resizebox{1.\textwidth}{!} {
\begin{tabular}{lrrrr}
\hline
\textbf{Description} & \textbf{\# (10 shards)} & \textbf{\%} & \textbf{\# (1024 shards estimate)} &  \\ \hline
Total examples & 1,652,725 & 100 & 169,239,040 &  \\
At least one "bad example filter" & 361,732 & 21.89 & 37,041,357 &  \\
Less than 200 unique tokens & 269,893 & 16.33 & 27,637,043 &  \\
Number of words outside range of 50 to 100,000 & 170,928 & 10.34 & 17,503,027 &  \\
Less than 50 words & 170,928 & 10.34 & 17,503,027 &  \\
Less than 80\% of words containing a alphabetic character & 80,836 & 4.89 & 8,277,606 &  \\
More than 30\% of lines ending with an ellipsis sign ("...") & 28,289 & 1.71 & 2,896,794 &  \\
Mean word length outside range of 3 to 10 characters & 22,280 & 1.35 & 2,281,472 &  \\
Less than 2 \texttt{nltk} Portuguese stopwords & 18,616 & 1.13 & 1,906,278 &  \\
Symbol ratio ("..." or "\#") greater than 0.1 & 8,805 & 0.53 & 901,632 &  \\
More than 90\% of lines starting with a bullet sign ("*") & 16 & 0.00 & 1,638 &  \\
More than 100,000 words & 0 & 0.00 & 0 &  \\
\hline
\end{tabular}
}
\label{tab:mc4_filters}
\end{table}

\end{document}